\documentclass[letterpaper, 10pt, journal, twoside]{ieeetran}      

\IEEEoverridecommandlockouts                              

\usepackage{multicol}
\usepackage[fleqn]{amsmath}
\usepackage{amssymb}
\usepackage{comment}
\usepackage{color}
\usepackage{graphicx}
\usepackage{mathtools}
\usepackage{breqn}
\usepackage{makecell}
\usepackage{caption}
\usepackage{subcaption}
\usepackage{afterpage}
\usepackage{siunitx}
\usepackage[hidelinks,bookmarks=true]{hyperref}

\allowdisplaybreaks

\newcommand{\transpose}[1]{#1^\intercal}

\newcommand{\argmax}{\operatornamewithlimits{argmax}}
\def\argmax{\mathop{\rm argmax}}

\def\atan2{\mathop{\rm atan2}}

\title{DCT Maps: \\
Compact Differentiable Lidar Maps Based on the Cosine Transform
}

\author{Alexander Schaefer, Lukas Luft, Wolfram Burgard
	\thanks{\copyright\ 2017 IEEE. Personal use of this material is permitted.  Permission from IEEE must be obtained for all other uses, in any current or future media, including reprinting/republishing this material for advertising or promotional purposes, creating new collective works, for resale or redistribution to servers or lists, or reuse of any copyrighted component of this work in other works.}
	\thanks{Manuscript received: September 10, 2017; Revised December 1, 2017; Accepted December 27, 2017.}%Use only for final RAL version
	\thanks{This paper was recommended for publication by Editor Cyrill Stachniss upon evaluation of the Associate Editor and Reviewers' comments.}
	\thanks{This work was partially supported by the European Commission in the Horizon 2020 framework program under grant agreement 644227-Flourish.}
	\thanks{All authors are with the Department of Computer Science, University of Freiburg, Germany.}%
	\thanks{\tt \small \{aschaef,luft,burgard\}@cs.uni-freiburg.de}
	\thanks{Digital Object Identifier (DOI): see top of this page.}
}%

\begin{document}

\maketitle

\begin{abstract}
Most robot mapping techniques for lidar sensors tessellate the environment into pixels or voxels and assume uniformity of the environment within them. 
Although intuitive, this representation entails disadvantages: 
The resulting grid maps exhibit aliasing effects and are not differentiable.
In the present paper, we address these drawbacks by introducing a novel mapping technique that does neither rely on tessellation nor on the assumption of piecewise uniformity of the space, without increasing memory requirements.
Instead of representing the map in the position domain, we store the map parameters in the discrete frequency domain and leverage the continuous extension of the inverse discrete cosine transform to convert them to a continuously differentiable scalar field in the position domain, which we call DCT map. 
A DCT map assigns to each point in space a lidar decay rate, which models the local permeability of the space for laser rays. 
In this way, the map can describe objects of different laser permeabilities, from completely opaque to completely transparent. 
DCT maps represent lidar measurements significantly more accurate than grid maps, Gaussian process occupancy maps, and Hilbert maps, all with the same memory requirements, as demonstrated in our real-world experiments. 
\end{abstract}

% Keywords appear just beneath the abstract. Use only for final RAL version. 

\section{Introduction}
\label{sec:introduction}

\begin{figure}
\centering
\begin{subfigure}{\linewidth}
	\centering
	\includegraphics[width=0.708\linewidth]{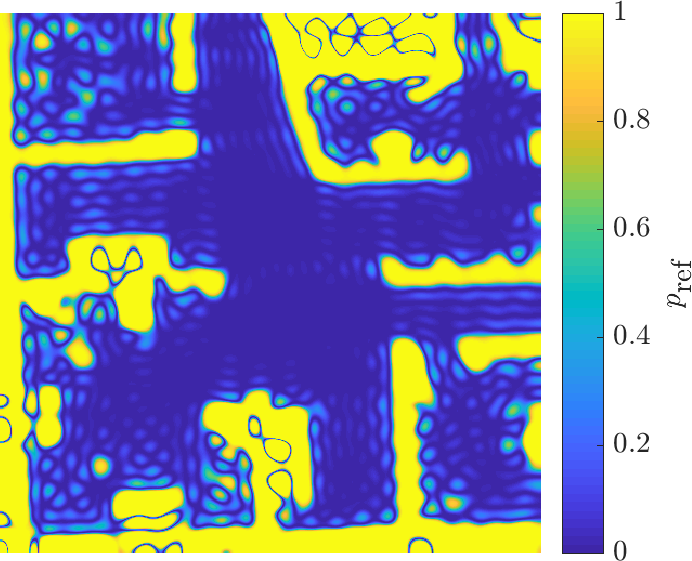}
	\caption{DCT map with $40 \times 40$ parameters.}
	\label{fig:dctmap}
\end{subfigure}
\begin{subfigure}{\linewidth}
	\centering
	\includegraphics[width=0.708\linewidth]{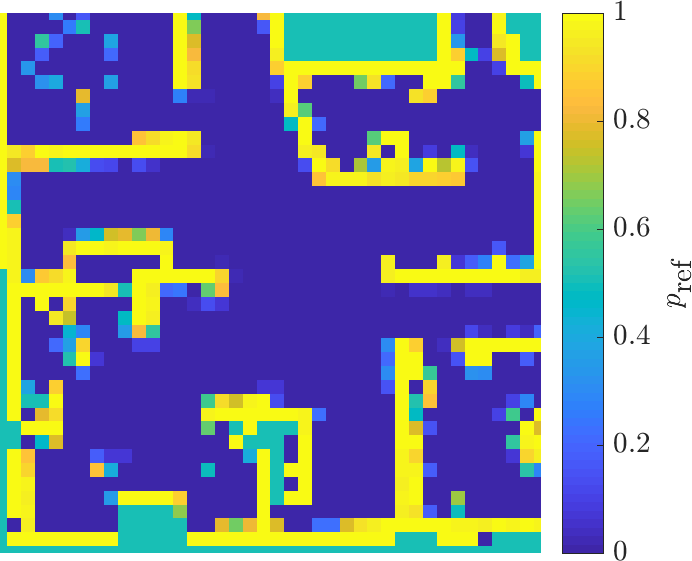}
	\caption{Grid map composed of $40 \times 40$ pixels with edge length \SI{25}{cm}.}
	\label{fig:gmmap}
\end{subfigure}
\begin{subfigure}{\linewidth}
	\centering
	\includegraphics[width=0.708\linewidth]{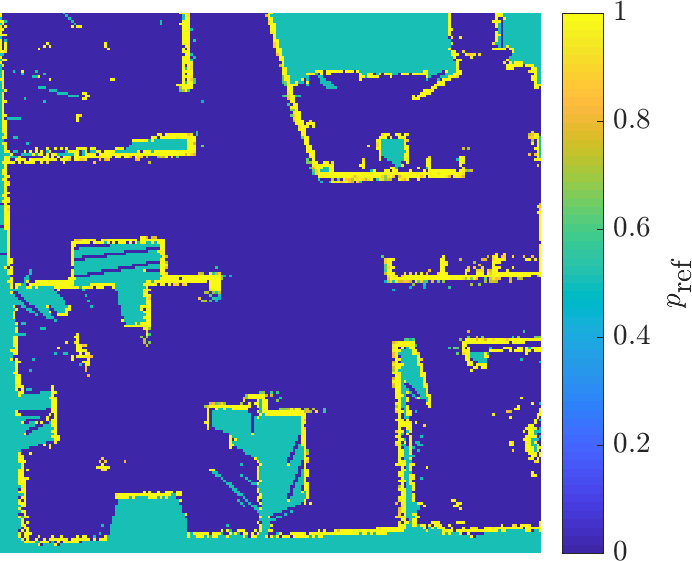}
	\caption{Grid map composed of $200 \times 200$ pixels with edge length \SI{5}{cm}.}
	\label{fig:gtmap}
\end{subfigure}
\caption{Decay rate maps of the same \mbox{$\SI{10}{m} \times \SI{10}{m}$} patch of the Intel Research Lab dataset \cite{howard2003} generated from the identical set of planar lidar measurements. 
The colors encode the reflection probability $p_{\textrm{ref}} \coloneqq 1-\exp(-\lambda)$, where $\lambda$ denotes the local laser decay rate.}
\label{fig:maps}
\end{figure}

\IEEEPARstart{M}{apping} and localization are at the heart of almost every mobile robotic system. 
In this context, lidar is a popular sensor modality, as lidar sensors produce relatively accurate, low-noise signals. 
Using these signals for mapping and localization requires an inverse and a forward sensor model. 
The inverse sensor model converts recorded measurements to a map. 
The forward model uses this map to assess the probability of incoming sensor readings given the robot pose.
The maps produced by the inverse pass are often grid maps: 
They tessellate the physical space into square pixels or cubic voxels.
Each pixel or voxel contains a value that is assumed to be constant within it. This value characterizes the statistical optical properties of the corresponding portion of space. 
Fig.~\ref{fig:gmmap} shows an example of such a grid map built from 2-D lidar scans recorded in an office environment.

Although tessellation is intuitive, grid maps bring with them several drawbacks. 
First, they can only coarsely approximate the true spatial distribution of the optical properties of interest. 
Aliasing effects occur whenever the optical characteristics of the environment change, as these transitions are never perfectly aligned with the raster of the grid. 
The grid map in fig.~\ref{fig:gmmap} exhibits the resulting characteristic staircase patterns. 
Although increasing the map resolution can theoretically alleviate this problem (see fig.~\ref{fig:gtmap}), quadratic or cubic memory complexity quickly renders this approach prohibitive. 
Depending on the use case, non-cubic voxels may mitigate the errors induced by tessellation~\cite{ryde2009b}.
Second, grid maps are not continuously differentiable, although this is a desirable property of any map. 
Continuous differentiability would allow to localize the robot by maximizing the measurement likelihood over the robot poses, and even to perform SLAM by maximizing the measurement likelihood over the robot poses and the map parameters.

In the present paper, we choose a different approach to avoid the aforementioned detrimental effects without increasing the memory demands.
Inspired by well-established digital image compression algorithms like JPEG, we store the map parameters in the discrete frequency domain and use the so-called continuous extension of the inverse discrete cosine transform~\cite{atoyan2004} to obtain a continuously differentiable scalar field in the position domain. 
In addition to the regular inverse discrete cosine transform, its continuous extension not only computes the function values at discrete grid points in the spatial domain, but also closely approximates them in between. 
We combine this map representation with the recently developed decay rate model for lidar sensors~\cite{schaefer2017}. 
The resulting DCT maps model the local permeability of the space for laser rays.
Fig.~\ref{fig:dctmap} depicts such a DCT map.
It was built from the identical information as the grid map in fig.~\ref{fig:gmmap} and has the same memory footprint, but it does not exhibit staircase patterns and better preserves the map contours.
Indeed, our experiments show that DCT maps represent lidar data with higher accuracy than other approaches.
Moreover, the continuous derivatives of DCT maps can be calculated in closed form, a fact that enables optimization-based SLAM.

In the following, we first survey different map representations. 
Then, we describe the mathematics behind DCT maps in detail. 
Finally, the findings of experiments conducted with publicly available real-world 2-D lidar datasets are presented.

\section{Related Work}
\label{sec:related_work}

Occupancy grid maps~\cite{moravec1985} were among the first probabilistic map representations used in robotics and are still widely used today. 
They tessellate the space into independent cells and assign each cell the posterior probability of being occupied. 
Occupancy grid maps cannot model semi-transparent objects; they assume that each cell is either completely free or completely occupied. 
In contrast, the decay rate model, which we employ to formulate DCT maps, explicitly models the permeability of each cell for a laser ray. 
If used in conjunction with grid maps, it even allows to calculate posterior distributions over the decay rate values without additional computational effort~\cite{luft2017}.

While 2-D occupancy maps are able to model large areas, even moderately-sized 3-D occupancy grids quickly outgrow the memory limitations of modern computers. 
For this reason, several research projects target compressed map representations.
Elevation maps~\cite{herbert1989} assume that the environment can be represented by a 2-D grid map whose cells contain not only occupancy values, but also one height coordinate per cell. 
To relax the assumption that the world is a single surface, \cite{triebel2006} extends elevation maps to multi-level surface maps.
Multi-volume occupancy grids~\cite{dryanovski2010} manage volumetric data as 2-D arrays, too, but in contrast to multi-level surface maps, each cell contains a list of occupied height regions and one of free height regions. 
Octrees~\cite{meagher1982} approach the memory limitation problem by hierarchically partitioning the space using an octal tree data structure. 
They have found broad application in robotics to model the spatial distribution of the occupancy value~\cite{wurm2010,payeur1997,pathak2007}. 
The authors of~\cite{fairfield2007} present an octree-based data structure that is efficient to update and to copy, so it can be used in particle filter-based SLAM, where hundreds of maps must be maintained in parallel. 
To model the dynamics of the environment, \cite{krajnik2014} assumes that the occupancy values in an octree are subject to periodic changes.
For each cell, the authors record the occupancy value over time and transform the resulting function to the frequency domain to predict the occupancy value at a later point in time. 
In this way, they achieve high compression ratios compared to storing one occupancy map per time step.
Multi-resolution occupied voxel lists~\cite{ryde2009a} differ from traditional occupancy mapping in that they store only the positions of the voxels that have been observed more frequently as occupied than as free. 
They are neither able to differentiate between unoccupied and unknown volumes, nor to account for semi-transparent voxels. 

The normal distributions transform~\cite{biber2003} was initially conceived in the context of scan matching. 
Based on this work, \cite{saarinen2013a} introduces the so-called normal distributions transform occupancy map. 
Basically, this map is a grid map, but instead of a single scalar, every cell contains a normally distributed occupancy probability density function, which is cropped at the voxel bounds. 
In this way, it drops the assumption that the space is uniform within each voxel. 
As opposed to DCT maps, however, normal distributions occupancy maps achieve higher accuracy at the cost of increased memory consumption.
Like all other occupancy-based approaches, they are not able to model semi-permeable objects, either.
Normal distributions occupancy maps are extended and advanced in \cite{saarinen2013b, stoyanov2013, valencia2014}.

Other approaches completely abandon the notion of voxels. 
For example, \cite{yguel2007} uses Haar wavelets to represent \mbox{3-D} occupancy data. 
The authors of~\cite{fridovich2017} drop the restriction that the elementary volumes of a map shall fill the space without gaps. 
Instead, they model the environment by non-overlapping spheres of equal sizes. 
In this way, they are able to more closely represent curved surfaces. 

Point clouds are a simple and convenient way to represent lidar sensor data.
However, in contrast to occupancy maps or decay rate maps, they are lossy in the sense that they store only the endpoints of the rays. 
They discard the ray path information of both reflected rays and rays that are not reflected. 
When point clouds are used for mapping, they accumulate memory for every incoming measurement, which limits their suitability for long-term navigation. 
Despite their drawbacks, many SLAM systems~\cite{nuechter2007,surmann2003,cole2006} represent lidar data in the form of point clouds.

In object reconstruction in computer graphics, objects are modeled as line segments in 2-D~\cite{chatila1985} or as polygon meshes in 3-D~\cite{montemerlo2002}. 
The resulting models can achieve an astonishing level of detail~\cite{levoy2000}. 
However, similar to mapping approaches based on implicit shape potentials like KinectFusion~\cite{newcombe2011}, they are not perfectly suited for lidar-based robot localization due to their sheer memory footprint and their inability to deal with semi-transparent materials.

Recently, machine learning techniques have completely relaxed the independence assumption between grid cells and produce continuous occupancy maps.
Gaussian process occupancy maps~(GPOM), for example, learn the environment of a robot and predict future states~\cite{ocallaghan2009,ocallaghan2012}.
Building on the latter, the authors of \cite{jadidi2017b} present an incremental GPOM formulation that enables online mapping.
Gaussian processes have also been applied to other map representations like implicit shape potentials~\cite{dragiev2011}.
Hilbert maps~\cite{ramos2016} are continuous occupancy maps built by projecting the lidar measurements in a Hilbert space, learning a logistic regression classifier, and then classifying each point in space as free or occupied.

\section{Approach}
\label{sec:approach}

In this section, we shortly revisit how the decay-rate model computes measurement probabilities conditioned on any kind of map, then we define the map using the continuous extension of the inverse discrete cosine transform. 
With these prerequisites, we derive the forward model, which computes the probability of a lidar measurement given the spectral parameters of the DCT map. 
In the last step, we formulate the inverse model as an optimization problem: 
We estimate the map parameters by maximizing the joint likelihood of all measurements collected during mapping. 

For brevity and without loss of generality, the following derivation is performed for \mbox{2-D} space. 
The derivation of the forward and inverse sensor model in \mbox{3-D} exactly parallels the \mbox{2-D} case.

\subsection{The Decay Rate Model}
\label{subsec:model}

The decay-rate model \cite{schaefer2017} models the probability that a lidar ray traverses a uniform medium as exponential decay process. 
The corresponding map assigns a decay rate to each point in space. 
This decay rate is a non-negative real number that describes the interaction between the laser ray and the environment completely.

To formulate the forward model mathematically, we introduce the following definitions. 
A lidar measurement \mbox{$z \coloneqq \{s,v,r\}$} describes a ray that originates at the sensor position $s$, travels in direction $v$, and ends after having traveled distance $r$. 
Assuming that the sensor provides its true position $s$, the true ray direction $v$, and that we are given a specific map~$\mathcal{M}$, we model the non-deterministic interaction between the ray and the environment by the measurement probability density with respect to the radius
\begin{align} 
\label{eq:p}
p(r) \coloneqq p(r \mid s, v, \mathcal{M}).
\end{align} 
Consequently, the absolute probability that the ray covers at least distance~$r$ is
\begin{align}
\label{eq:N}
\mathcal{N}(r) \coloneqq 1-\int_0^r p(r') \, dr'.
\end{align}
Alternatively, we can express equation~\eqref{eq:N} in form of the differential equation
\begin{align}
\label{eq:pN}
p(r) = -\frac{d\mathcal{N}(r)}{dr}.
\end{align}
The essential idea of the decay rate model consists in the assumption that $\mathcal{N}(r)$ obeys an exponential decay process
\begin{align}
\label{eq:differential_equation}
\frac{d\mathcal{N}(r)}{dr} 
	= -\lambda\left(r\right) \, \mathcal{N}\left(r\right),
\end{align}
where $\lambda(r)$ denotes the decay rate at a specific radius $r$ along the ray.
By combining this model assumption with differential equation~\eqref{eq:pN}, we obtain the measurement probability density
\begin{align}
\label{eq:plN}
p(r) = \lambda(r) \, \mathcal{N}(r).
\end{align}
In \eqref{eq:differential_equation} and \eqref{eq:plN}, $\lambda(r)$ is obtained by evaluating the map $\lambda(x,y)$ along the trajectory of the ray.

The above formulation of the decay rate model is independent of any specific map representation. 
To use it as forward model in combination with DCT maps, we need to define the map function~$\lambda(r)$ and solve the differential equation. 
In order to do so, we describe the spatial representation of DCT maps in the next section in detail.
After that, we have all prerequisites at hand to solve the differential equation.
The solution enables us to express the measurement probability of a lidar measurement given the map in closed form.

\subsection{Transforming the Spectral Map Representation to the Spatial Domain}
\label{subsec:ceidct}

To avoid the disadvantages related to tessellation, DCT maps represent the map parameters in the discrete frequency domain instead of the position domain. 
Calculating the measurement likelihood from such a representation requires the definition of the transformation from the frequency domain to the spatial domain. 
We employ the continuous extension of the inverse discrete cosine transform (CEIDCT)~\cite{atoyan2004}. 
Like other continuous extensions of Fourier-related transforms, it converts a discrete signal in the frequency domain to a continuous signal in the spatial domain. 
However, it differs from its relatives in that the continuous signal converges to the continuous function from which it was sampled for an increasing number of parameters (see \cite{atoyan2004}, pp.~11--12).
Moreover, its parameters are purely real-valued.
For these reasons, it is particularly suited for our use case.

If we assume the spectral map parameters to be given by a matrix $\mathcal{A}$ with $L$ rows and $M$ columns, and if we denote the elements of $\mathcal{A}$ by $a_{lm}$ with $l \in \{0,1,\ldots,L-1\}$ and $m \in \{0,1,\ldots,M-1\}$, the CEIDCT transforms them to a continuously differentiable decay rate map defined for each point $(x,y)$ in the spatial domain
\begin{align}
\lambda(x,y) 
	&= \left(
			\sum_{l=0}^{L-1} \sum_{m=0}^{M-1} a_{lm}
				\cos\left(l\tilde{x}\right) \cos\left(m\tilde{y}\right)
		\right)^2 
		\label{eq:double_sum} 
		\\
	&= \left( 
			\sum_{i=0}^{I-1} a_i 
				\cos(l_i \tilde{x}) \cos(m_i \tilde{y}) 
		\right)^2 \notag \\
\begin{split}	
	&= \sum_{i=0}^{I-1} \sum_{j=0}^{I-1}
		a_i \cos\left(l_i \tilde{x}\right)
					\cos\left(m_i \tilde{y}\right) \\
	&\phantom{= \sum_{i=0}^{I-1} \sum_{j=0}^{I-1}}
		a_j \cos\left(l_j \tilde{x}\right)
					\cos\left(m_j \tilde{y}\right)
\end{split} \notag \\
\begin{split}
	&= \frac{1}{8} \sum_{i=0}^{I-1} \sum_{j=0}^{I-1} a_i \, a_j	
		\sum_{\alpha \in Q} \sum_{\beta \in Q} \sum_{\gamma \in Q} \\
	&\phantom{= \frac{1}{8}} \cos
	\Big(
		(l_i+\alpha l_j) \tilde{x} 
		+ \beta(m_i+\gamma m_j) \tilde{y}
	\Big)
\end{split}
\label{eq:spatial_lambda}
\end{align}
with \mbox{$I \coloneqq L\,M$} and \mbox{$Q \coloneqq \{-1,+1\}$}.
The tildes denote the $\pi$-normalization of the map coordinates: $\tilde{x} \coloneqq \frac{\pi x}{X}$, $\tilde{y} \coloneqq \frac{\pi y}{Y}$, where $X$ and $Y$ indicate the extent of the map. The variables $l_i$ and $m_i$ are the row and column indices into the matrix $\mathcal{A}$ that correspond to its $i$-th element~$a_i$.

The original formulation of the CEIDCT does not square the double sum in \eqref{eq:double_sum}. 
We employ this variant, however, because it ensures that the decay rate is non-negative for every point in the map. 
Negative decay rates would cause problems, as we cannot interpret the negative measurement probabilities in which they might result.

To solve equation~\eqref{eq:differential_equation}, we still need to transition from $\lambda(x,y)$ to $\lambda(r) \coloneqq \lambda(r,s,v)$. To that end, we apply the ray equation \mbox{$\transpose{[x,y]} = s + v \, r$} to \eqref{eq:spatial_lambda} and obtain
\begin{align}
\label{eq:lambda_ray}
\begin{split}
\lambda(r)
	&= \frac{1}{8} \sum_{i=0}^{I-1} \sum_{j=0}^{I-1}  a_i \, a_j 	
		\sum_{\alpha \in Q} \sum_{\beta \in Q} \sum_{\gamma \in Q} \\
	&\phantom{= \frac{1}{8}} \cos
	\Big(
		(l_i+\alpha l_j)\left[\tilde{s}_x + \tilde{v}_x r\right] \\
	&\phantom{= \frac{1}{8} \cos \Big(}
		+ \beta(m_i+\gamma m_j)\left[\tilde{s}_y + \tilde{v}_y r\right]
	\Big).
\end{split}
\end{align}

\subsection{Computing the Measurement Likelihood}
\label{subsec:measurement_likelihood}

Now we express the measurement probability of a lidar ray as a function of the measurement $z$ and the spectral representation of the map $\mathcal{A}$ by solving the differential equation \eqref{eq:differential_equation}. The solution is 
\begin{equation}
\label{eq:differential_equation_solution}
\mathcal{N}(r) 
	= \exp\big\{ -\mathcal{S}\left( s,v,r \right) \big\}
\end{equation}
with
\begin{dmath*}
\label{eq:S}
\mathcal{S}\left(s, v, r\right) 
	= \int_0^r \lambda(r') \, dr' \\
	= \frac{1}{8} \sum_{i=0}^{I-1} \sum_{j=0}^{I-1} a_i \, a_j
		\sum_{\alpha \in Q} \sum_{\beta \in Q} \sum_{\gamma \in Q} 
		A_{ij}
\end{dmath*}
where
\begin{dmath}
A_{ij} {:=} A(i, j, \alpha, \beta, \gamma) \\
=
\begin{cases}
\frac{
	\left[
		\sin\left((l_i+\alpha l_j)
			\left[\tilde{s}_x + \tilde{v}_x r'\right] 
				+ \beta(m_i+\gamma m_j)\left[\tilde{s}_y 
				+ \tilde{v}_y r'\right]\right)
	\right]_0^r}
{(l_i+\alpha l_j)\tilde{v}_x + \beta(m_i+\gamma m_j)\tilde{v}_y}, \\
	\qquad \textrm{if } 
		(l_i+\alpha l_j)\tilde{v}_x + \beta(m_i+\gamma m_j)\tilde{v}_y \neq 0
\\
\\
r \cos\left( \left( l_i + \alpha l_j \right) \tilde{s}_x 
		+ \beta \left( m_i + \gamma m_j \right) \tilde{s}_y \right), \\
	\qquad \textrm{if } 
		(l_i+\alpha l_j)\tilde{v}_x + \beta(m_i+\gamma m_j)\tilde{v}_y = 0
\end{cases}
\label{eq:A}
\end{dmath}
Note that out of the infinite number of solutions to \eqref{eq:differential_equation}, we chose the one that satisfies the boundary condition \mbox{$\mathcal{N}(0) \stackrel{!}{=} 1$}.

By plugging equations \eqref{eq:lambda_ray} and \eqref{eq:differential_equation_solution} in \eqref{eq:plN}, we finally obtain the closed-form solution of the measurement likelihood $p(r)$ for rays with real-valued radius $r$. 

Not all lidar measurements are real-valued, though. 
In practice, the range of every lidar scanner is limited to a finite interval \mbox{$\left[r_{\min}, r_{\max}\right]$}.
We call the rays reflected outside this interval no-return rays.
In the following, we assume that the sensor identifies rays falling below $r_{\min}$ by the tag $\mathrm{sub}$ and rays that exceed $r_{\max}$ by the tag $\mathrm{super}$. 
Consequently, the space of all possible measurements $r$ is the mixed discrete-continuous set \mbox{$\mathcal{D} \coloneqq \{\mathrm{sub}, \mathrm{super}, r': r' \in [r_{\mathrm{min}}, r_{\mathrm{max}}]\}$}.

Fortunately, the decay-rate model easily accommodates both sorts of no-return rays:
\begin{align}
P(\mathrm{sub}) 
	&= \int_0^{r_{\min}} p(r) \, dr 
	= 1 - \mathcal{N}(r_{\min}), \label{eq:Psub} \\
P(\mathrm{super})
	&= \int_{r_{\max}}^{\infty} p(r) \, dr
	= \mathcal{N}(r_{\max}). \label{eq:Psuper}
\end{align}
Supporting no-return rays is an important feature of the model. In a typical outdoor setting, no-return rays represent a considerable fraction of all measurements. If the model is unable to incorporate the information they convey, which is the case for the endpoint model, for example, it will inevitably loose accuracy.

During mapping and localization, one does not need to evaluate the measurement probability of a single ray, but of a whole laser scan \mbox{$Z \coloneqq \{z_k\}$} consisting of $K$ rays, both with real-valued radius and without detected reflection.
To obtain this probability, we first formulate the probability density function for each ray over the mixed space $\mathcal{D}$ by combining equations \eqref{eq:plN}, \eqref{eq:Psub}, and \eqref{eq:Psuper} to
\begin{align*}
p\left(z \mid \mathcal{M}\right)
	&\coloneqq 
		\begin{cases}
			P(\mathrm{sub}), 
				&\textrm{if } r = \mathrm{sub} \\
			p(r),	
				&\textrm{if } r \in \left[r_\min, r_\max\right] \\
			P(\mathrm{super}),
				&\textrm{if } r = \mathrm{super}
		\end{cases}
\end{align*}
The above result, which we call a mixed probability density, is positive, real, and integrates to unity.
Using the independence assumption, we then compute the joint probability density of all rays as the product 
\begin{align*}
p\left(Z \mid \mathcal{M}\right) 
= \prod_{k=0}^{K-1} p\left(z_k \mid \mathcal{M}\right).
\end{align*}

\subsection{Building the Decay Rate Map}
\label{subsec:mapping}

During the inverse pass, we want to determine the map parameters $\mathcal{A}$ that best explain the lidar measurements collected in the mapping run:
\begin{align*}
\mathcal{A} 
	= \argmax_{\mathcal{A}} p(Z \mid \mathcal{A})
	= \argmax_{\mathcal{A}} \log\Big\{p(Z \mid \mathcal{A})\Big\}.
\label{eq:optimization}
\end{align*}
This non-linear multivariate optimization problem can be solved by iterative computational optimization techniques like stochastic gradient descent or trust-region methods. 
These methods work considerably more reliable and faster when provided with first-order and second-order analytical logarithmic derivatives with respect to the spectral map parameters. 
To calculate the derivatives, we introduce the following shortcut notation:
\begin{align*}
\frac{\partial \lambda(x,y)}{\partial a_i} 
	\eqqcolon \sum_{j=0}^{I-1} a_j \, B_{ij} 
	\eqqcolon B_i, 
\end{align*}
with
\begin{align*}
B_{ij} \coloneqq 2 \cos(l_i \tilde{x}) \cos(m_i \tilde{y}) \cos(l_j \tilde{x}) \cos(m_j \tilde{y})
\end{align*}
and
\begin{align*}
\frac{\partial \mathcal{N}}{\partial a_i}
&= -\mathcal{N} \,
	\frac{\partial \mathcal{S}\left(s,v,r\right)}{\partial a_i} 
	\eqqcolon -\mathcal{N} \, C_i
\end{align*}
where
\begin{align*}
C_i
&= \frac{1}{8} \sum_{j=0}^{I-1} a_j
	\sum_{\alpha \in Q} \sum_{\beta \in Q} \sum_{\gamma \in Q} A_{ij} + A_{ji}
		\eqqcolon \sum_{j=0}^{I-1} a_j \, C_{ij}
\end{align*}
with $A_{ij}$ as defined in \eqref{eq:A}.
Using this notation, we can express the first-order logarithmic derivative of the absolute probability \mbox{$P(z \mid \mathcal{A})$} of a single measurement in a compact way:
\begin{equation*}
\frac{\partial \log\left\{p(z \mid \mathcal{A})\right\}}{\partial a_i}
= 
\begin{cases}
\frac{\mathcal{N} C_i}{1-\mathcal{N}},
	& \textrm{if } r = \mathrm{sub} \\
\frac{B_i}{\lambda} - C_i,
	& \textrm{if } r \in \left[ r_\min, r_\max \right] \\
-C_i,
	& \textrm{if } r = \mathrm{super}
\end{cases}
\end{equation*}
The derivative of the joint absolute measurement probability is then simply the sum of the derivatives of the individual measurement likelihoods
\begin{equation*}
\frac{\partial \log\left\{p(Z \mid \mathcal{A})\right\}}{\partial a_i}  
	= \sum_{k=0}^{K-1} \frac{\partial \log\left\{p(z \mid \mathcal{A})\right\}}{\partial a_i}.
\end{equation*}

The second-order derivatives of the measurement likelihood with respect to the map parameters are given by 
\begin{dmath*}
\frac{\partial^2 \log\left\{p(z \mid \mathcal{A})\right\}}
	{\partial a_j \partial a_i} \notag 
= 
\begin{cases}
\frac{N \left( C_{ij} - C_i C_j \right)}{1-N} 
+ \frac{N^2 C_i C_j}{\left( 1-N \right)^2},
	& \textrm{if } r = \mathrm{sub} \\
\frac{B_{ij}}{\lambda} - \frac{B_i B_j}{\lambda^2} - C_{ij},
	& \textrm{if } r \in \left[r_\min, r_\max\right] \\
-C_{ij},
	& \textrm{if } r = \mathrm{super}
\end{cases}
\end{dmath*}

As before, the second-order derivative of the joint measurement log-likelihood is the sum of the second-order derivatives of all measurements.

\section{Experiments}

To assess how well DCT maps represent lidar data in comparison to existing mapping approaches, we conduct three series of experiments. 
In the first series, we compare the spatial map values of DCT maps and grid maps with identical memory requirements to a ground truth map and use the resulting error as a measure of map accuracy.
In the second series, we evaluate the likelihoods that DCT maps, grid maps, Gaussian process occupancy maps, and Hilbert maps assign to measurements that were used to build them. 
The higher this likelihood, the better the respective map explains the underlying data.
We conclude this section with a comparison of the empirical execution times of the different approaches.

The data at the basis of our experiments stems from rich planar lidar datasets recorded in spacious indoor environments.
Each set contains the corresponding robot poses computed by SLAM, which we use as ground truth poses to build all maps.
The data is publicly available from the Robotics Data Set Repository~\cite{howard2003}.
Table~\ref{tab:datasets} shows which datasets were used in our experiments.

\begin{table}[p]
\def\arraystretch{1.2}
\centering
\begin{tabular}{l | l | l}
Acapulco Conv. Ctr. & Edmonton Conv. Ctr.	   & Uni. Freiburg, 101 \\
Uni. Texas, ACES3   & FHW museum			   & Infinite corridors \\ 
Belgioioso Castle   & Uni. Washington, Seattle & Intel Research Lab \\
MIT, CSAIL          & Uni. Freiburg, 079 	   & \"Orebro University \\
\end{tabular}
\caption{The 12 datasets taken from the Robotics Data Set Repository~\cite{howard2003} and used in all three experiment series.}
\label{tab:datasets}
\end{table}

\subsection{Map Value Comparison}
\label{subsec:mapcomparison}

In this experiment series, we compare the map values of DCT maps and decay rate grid maps of different resolutions to the values of a decay rate ground truth grid map. 
All grid maps are computed according to the algorithm described in \cite{schaefer2017}.
At the beginning, for each dataset, we create a fine-grained ground truth grid map that covers a $\SI{10}{m} \times \SI{10}{m}$ patch densely filled with $10^4$ lidar measurements.
It consists of $200 \times 200$ pixels with an edge length of \SI{0.05}{m}. 
Then, we use the same sets of measurements to build pairs of one DCT map and one grid map, respectively, for each dataset and each resolution. 
The maps in these pairs possess the same number of parameters and require the identical amount of memory. 
We use five different map resolutions: \num{10x10}, \num{13x13}, \num{20x20}, \num{29x29}, and \num{40x40}.
For grid maps, they correspond to pixel edge lengths of \SI{1.00}{m}, \SI{0.75}{m}, \SI{0.50}{m}, \SI{0.35}{m}, and \SI{0.25}{m}.
To give an intuition of what these maps look like, fig.~\ref{fig:maps} exemplarily shows the $40 \times 40$ DCT map of the Intel Research Lab dataset, the corresponding grid map, and the ground truth grid map.

Having created the maps, we sample the ground truth map at the midpoints of all cells that were observed at least once and look up the corresponding values in the DCT map and in the grid map.
The resulting map values are hard to compare:
As the decay rate~$\lambda$ is defined over the half-open interval $[0,\infty)$, the map values might be infinite. 
In order to make them comparable, we employ the strictly increasing monotonic transformation function \mbox{$p_{\textrm{ref}}=1-\exp(-\lambda)$}, which maps every decay value to a finite value \mbox{$p_{\textrm{ref}} \in [0,1]$}. 
The value~$p_{\textrm{ref}}$ can be interpreted as the absolute probability that a ray is reflected before having traveled a distance of \SI{1}{m} in a hypothetical homogeneous medium of decay rate~$\lambda$.
Please note that the distance~\SI{1}{m} is an arbitrarily chosen parameter.
However, while surveying different distance values, we found out that varying this parameter has little effect on the quality of the results.
We compute the root mean squared error (RMSE) in $p_{\textrm{ref}}$ between the DCT maps and the ground truth map and between the grid maps and the ground truth map. 
Table~\ref{tab:mapcomp} condenses the corresponding results by determining the mean and the standard deviation of the RMSE values.
Additionally, it indicates the $p$-values of the one-sided paired-sample $t$-test. 
Small $p$-values indicate that the null hypothesis is unlikely and that the alternative hypothesis -- the mean RMSE of DCT maps is smaller than the one of grid maps -- becomes likely.

\begin{table}[p]
\def\arraystretch{1.2}
\centering
\begin{tabular}{c | c c | c c | r | c}
\multicolumn{1}{c}{} & \multicolumn{2}{c}{DCT} & \multicolumn{2}{c}{GM} & \multicolumn{2}{c}{} \\ 
$l\ [\si{m}]$ & $\mu$ & $\sigma$ & $\mu$ & $\sigma$ & \multicolumn{1}{c|}{$p$ [\si{\percent}]} & $\Delta_{\mu}$ [\si{\percent}] \\
\hline
1.00 & 0.3314 & 0.0679 & 0.3330 & 0.0708 & 39.36 & 0.48 \\
0.75 & 0.3146 & 0.0675 & 0.3319 & 0.0752 &  1.01 & 0.52 \\
0.50 & 0.2932 & 0.0645 & 0.3093 & 0.0690 &  0.03 & 5.21 \\
0.35 & 0.2571 & 0.0611 & 0.2822 & 0.0672 &  0.05 & 8.89 \\
0.25 & 0.2370 & 0.0563 & 0.2543 & 0.0583 &  0.07 & 6.80 \\
\end{tabular}
\caption{Mean and standard deviation of the absolute RMSE values of DCT maps and grid maps with respect to the ground truth map, computed over all datasets. 
GM denotes grid maps, $l$ is the pixel edge length, $\mu$ and $\sigma$ denote the mean and the standard deviation of the RMSE, respectively, and $p$ is the $p$-value of the one-sided paired-sample $t$-test. The variable $\Delta_{\mu} \coloneqq 1-\frac{\mu_{\textrm{DCT}}}{\mu_{\textrm{GM}}}$ indicates the improvement in RMSE of DCT maps relative to grid maps.}
\label{tab:mapcomp}
\end{table}

While at a resolution of $10 \times 10$, both map modalities hardly differ in terms of accuracy, for all finer resolutions, DCT maps outperform grid maps at a confidence level of at least \SI{99}{\percent}.
Note that the maximum gain in accuracy is located at a resolution of \num{29x29}; at \num{40x40}, DCT maps are still significantly more accurate than grid maps, but the gain is not as large as for \num{29x29}.
We attribute this to the fact that with increasing resolution, grid maps converge to the ground truth map, which itself is a grid map.

\subsection{Measurement Probability Comparison}
\label{subsec:probcomparison}

The maps computed in the first experiment series are maximum likelihood maps. 
Maximum likelihood maps shall maximize the measurement probability of the data that was used to create them. 
The higher this likelihood, the better the map represents the underlying data. 
Consequently, in the second experiment series, we interpret the likelihood a map assigns to its underlying data as a measure of its quality.
We compare four different approaches: DCT maps, decay rate grid maps, Gaussian process occupancy maps (GPOM), and Hilbert maps, which also model the spatial occupancy probability as a continuous scalar field.
More specifically, we use GPOMs with Mat\'ern kernel functions as described in \cite{jadidi2017b} and Hilbert maps with Nystr\"om features, which, according to \cite{ramos2016}, give the most accurate map results.
All hyperparameters are set as described in \cite{jadidi2017b} and \cite{ramos2016}, respectively.
The data at the basis of the experiments is the same as in the previous experiment series, but the number of lidar measurements is reduced to \num{500}.

The comparison is designed to guarantee that all maps have the same memory requirements in terms of numbers of real-valued parameters. 
For grid maps and DCT maps, we can ensure that by comparing maps with the same number of pixels and spectral parameters, respectively.
For GPOM, we randomly downsample the design matrix and the target vector so that the length of the Gaussian process parameter vector matches the number of grid pixels and spectral parameters, respectively.
For Hilbert maps, we set the number of components of the Nystr\"om features to the number of parameters corresponding to the specific resolution.

Now, we compute the joint measurement likelihood of all lidar measurements for each map modality.
For grid maps, we calculate the measurement likelihood according to \cite{schaefer2017}.
For DCT maps, we follow the equations given in section~\mbox{\ref{subsec:measurement_likelihood}}.
For GPOMs and Hilbert maps, we rasterize their continuous occupancy fields with a pixel edge length of \SI{0.05}{m}, perform ray tracing, and cumulate the occupancy probabilities along the rays.

\begin{table}[p]
	\def\arraystretch{1.2}
	\centering
	\begin{tabular}{c | c c | c c | c c}
		\multicolumn{1}{c}{} & \multicolumn{2}{c}{$lp_{\textrm{DCT}}-lp_{\textrm{GM}}$} & \multicolumn{2}{c}{$lp_{\textrm{DCT}}-lp_{\textrm{GPOM}}$} & \multicolumn{2}{c}{$lp_{\textrm{DCT}}-lp_{\textrm{HM}}$} \\
		$l\ [\si{m}]$ & $\mu$ & $\sigma$ & $\mu~[10^4]$ & $\sigma~[10^4]$ & $\mu~[10^4]$ & $\sigma~[10^4]$ \\
		\hline
		1.00 & \multicolumn{1}{r}{ 88.5} & \multicolumn{1}{r|}{ 74.1} & 4.21 & 3.00 & 4.30 & 3.42 \\
		0.75 & \multicolumn{1}{r}{150.6} & \multicolumn{1}{r|}{146.5} & 3.89 & 2.81 & 4.19 & 3.46 \\
		0.50 & \multicolumn{1}{r}{135.7} & \multicolumn{1}{r|}{ 63.9} & 3.53 & 2.60 & 4.06 & 3.42 \\
		0.35 & \multicolumn{1}{r}{196.1} & \multicolumn{1}{r|}{101.5} & 2.97 & 2.15 & 4.12 & 3.32 \\
		0.25 & \multicolumn{1}{r}{ 96.8} & \multicolumn{1}{r|}{132.0} & 2.68 & 1.90 & 4.12 & 3.31 \\
	\end{tabular}
	\caption{Mean and standard deviation of the log-likelihood differences between DCT maps and the other mapping approaches, computed over all datasets.
		The variable~$lp$ denotes the cumulated log-likelihood of all measurements in one dataset, GM denotes grid maps, HM means Hilbert maps, $l$ is the pixel edge length, and $\mu$ and $\sigma$ are the mean and the standard deviation of the log-likelihood differences, respectively.}
	\label{tab:pcomp}
\end{table}

Table~\ref{tab:pcomp} displays the resulting findings: the mean and standard deviation of the log-likelihood differences between DCT maps and the other approaches over all datasets.
After having performed Anderson-Darling tests to ensure that the measurement probability quotients are indeed log-normally distributed, we perform one-sided paired-sample $t$-tests.
For all resolutions, they indicate that DCT maps yield significantly higher measurement log-likelihoods at a confidence level of at least \SI{98.56}{\percent}.

\begin{figure*}
	\begin{subfigure}{0.1945\textwidth}
		\centering
		\includegraphics[width=\textwidth]{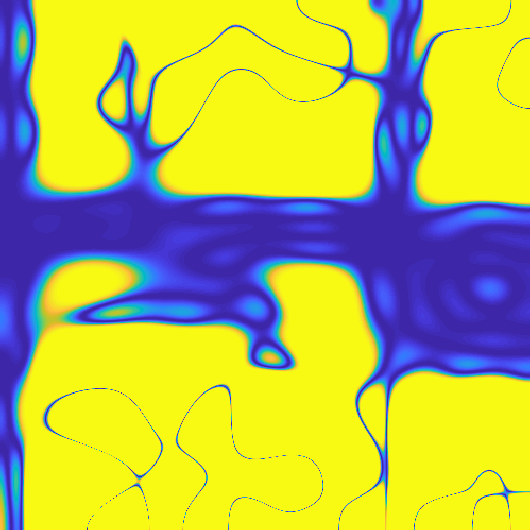}
		\caption{DCT map.}
		\label{fig:dct_coarse}
	\end{subfigure}
	\begin{subfigure}{0.1945\textwidth}
		\centering
		\includegraphics[width=\textwidth]{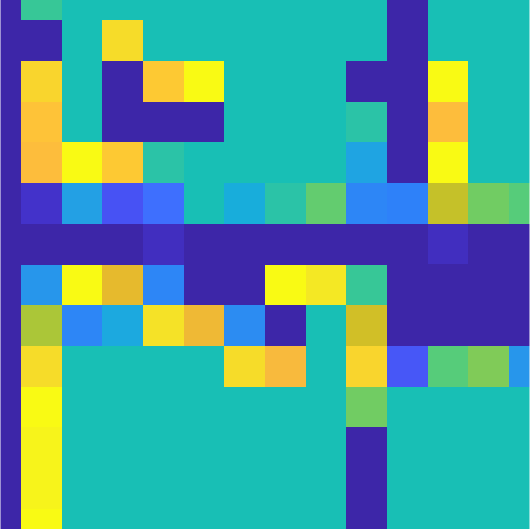}	
		\caption{Grid map.}
		\label{fig:gm_coarse}
	\end{subfigure}
	\begin{subfigure}{0.1945\textwidth}
		\centering
		\includegraphics[width=\textwidth]{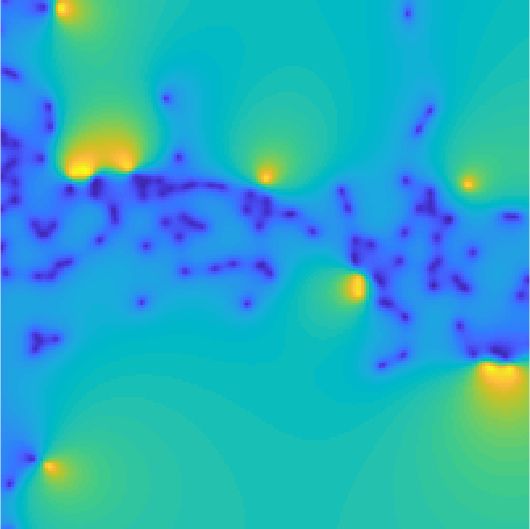}
		\caption{GPOM.}
		\label{fig:gpom_coarse}
	\end{subfigure}
	\begin{subfigure}{0.1945\textwidth}
		\centering
		\includegraphics[width=\textwidth]{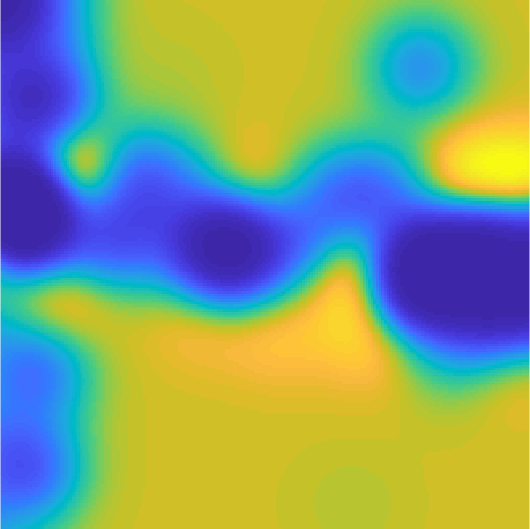}
		\caption{Hilbert map.}
		\label{fig:hm_coarse}
	\end{subfigure}
	\begin{subfigure}{0.1945\textwidth}
		\centering
		\includegraphics[width=\textwidth]{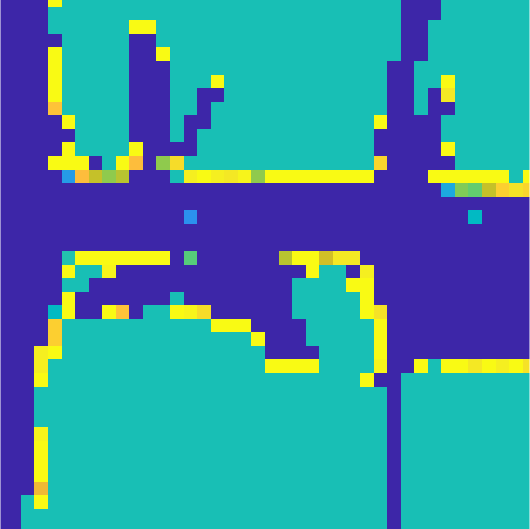}
		\caption{Ground truth grid map.}
		\label{fig:gt_coarse}
	\end{subfigure}
	\caption{Maps of different modalities created in the experiment series described in section~\ref{subsec:probcomparison} for the University of Washington dataset. 
	The four maps to the left all have the same memory requirements of a mere \num{169} real-valued parameters.
	The grid map (e) is given as ground truth with a resolution of \num{40x40}.
	The decay rate maps (a), (b), (e) show $p_{\textrm{ref}}$ values as described in section~\ref{subsec:mapcomparison}, the other maps show occupancy probabilities.
	Blue means 0, green means 0.5, yellow means 1.}
	\label{fig:mapmodalities}
\end{figure*}

The results show that the log-differences between DCT maps and grid maps are two magnitudes smaller than those between DCT maps and GPOM or Hilbert maps, respectively.
The level of the difference is influenced by the raster size chosen when computing the measurement likelihood for GPOMs and Hilbert maps.
But the main reason for these large differences is the fact that both GPOMs and Hilbert maps have comparatively high memory requirements.
GPOMs store the map information in the parameter vector.
The number of training points processed is proportional to the length of the parameter vector.
As we restricted this length, only a limited number of training points could be processed; as a consequence, the classification results of GPOMs remain rather vague.
As far as Hilbert maps are concerned, in \cite{ramos2016}, the authors recommend to use \num{1000} components for mapping with Nystr\"om features.
In our experiments, we use numbers as small as \num{100} parameters.
Additionally, both GPOMs and Hilbert maps suffer from the fact that they need to sample a limited number of free and occupied training points along the laser rays, whereas DCT maps and decay rate grid maps incorporate the full path information of an arbitrary number of rays.
Fig.~\ref{fig:mapmodalities} illustrates the resulting differences in accuracy between the maps produced by the four approaches for \num{13x13} parameters.

\subsection{Execution Times}

To give an intuition of the empirical runtime requirements of each of the methods used in the previous section, we average over ten mapping runs performed per method, dataset, and resolution.
Table~\ref{tab:exectimes} lists the medians of these averages over all datasets.
The measurements are collected on an Intel i7-2600K CPU running at \SI{3.40}{GHz}.
Grid maps, DCT maps, and GPOM are implemented in MATLAB R2017b.
The GPOM implementation is based on the publicly available GPML toolbox~\cite{rasmussen2010}.
To time Hilbert maps, we customized the Python implementation provided by \cite{ramos2016}.
The DCT optimization process is stopped once the relative change in the measurement log-likelihood is smaller than \num{1e-3}.

\begin{table}[tp]
	\def\arraystretch{1.2}
	\centering
	\begin{tabular}{c | r r r r}
		$l\ [\si{m}]$ &
			\multicolumn{1}{c}{$t_{\textrm{DCT}}~[\si{s}]$} &
			\multicolumn{1}{c}{$t_{\textrm{GM}}~[\si{s}]$} & \multicolumn{1}{c}{$t_{\textrm{GPOM}}~[\si{s}]$} & \multicolumn{1}{c}{$t_{\textrm{HM}}~[\si{s}]$} \\ 
		\hline
		1.00 &  3.52 & 0.0917 &  1.12 &  22.8 \\
		0.75 &  4.69 & 0.0915 &  1.98 &  40.6 \\ 
 		0.50 &  3.70 & 0.0923 &  3.25 & 108.4 \\ 
		0.35 & 18.25 & 0.0926 &  6.45 & 328.3 \\
		0.25 & 39.84 & 0.0942 & 14.04 & 949.3
	\end{tabular}
	\caption{
		Empirical execution time measurements collected during map creation.
		The variable~$t$ denotes the median of the mapping times over all datasets.
	}
	\label{tab:exectimes}
\end{table}

Table~\ref{tab:exectimes} indicates that grid maps are by far the fastest mapping technique.
DCT maps and GPOMs are approximately two orders of magnitude slower.
This is due to the fact that during the optimization phase, DCT maps and GPOMs need to consider all parameters, which leads to quadratic computational complexity in the number of parameters.
The most expensive operation in grid mapping, however, is ray casting, resulting in approximately linear computational complexity in the map resolution.
Hilbert maps are three to four magnitudes slower than grid maps, the reason for this probably being the non-differentiable nature of the objective function, which needs to be approximated using finite differences.

\section{Conclusion and Future Work}

We presented a novel map representation based on the recently introduced decay rate model for lidar sensors~\cite{schaefer2017}. 
In contrast to most conventional maps, our so-called DCT maps store the map parameters in the discrete frequency domain.
We applied the continuous extension of the inverse discrete cosine transform to the spectral parameters to obtain a continuously differentiable scalar field in the position domain.

Compared to other mapping approaches like decay rate grid maps, Gaussian process occupancy maps (GPOM), and Hilbert maps, the proposed approach results in significantly improved map accuracy, as demonstrated in extensive real-world experiments.
Moreover, DCT maps provide a ray tracing-based forward sensor model that allows to infer measurement probabilities directly from the spectral map representation in closed form, whereas the computation of ray tracing-based measurement probabilities based on continuous occupancy maps like GPOM and Hilbert maps necessitates the rasterization of the map and hence the introduction of a rasterization parameter.
As opposed to GPOM and Hilbert maps, DCT maps use the full ray path information when building the map instead of sampling points along the ray.

Due to the promising experimental results, we plan improvements and extensions of DCT maps.
First, we will address the issue that the computational complexity of DCT maps is higher than the one of grid maps, and that incremental updates require the repeated optimization of the map parameters.
More specifically, we will develop a hybrid approach that locally optimizes the map and that makes use of massive parallelization.
Second, we will extend the method by explicitly representing unexplored areas in the map.
Currently, DCT maps are not able to distinguish between observed and unobserved map regions.
Third, we will investigate how well DCT maps are suited for lossy compression.
Finally, we plan to present a complete SLAM system based on DCT maps in the near future.

\section*{Acknowledgments}

We thank Maani Ghaffari Jadidi for kindly supporting us with his GPOM implementation, Lionel Ott for releasing his Python implementation of Hilbert maps, and Patrick Beeson, Mike Bosse et~al., Dieter Fox, Dirk H{\"a}hnel, Nick Roy, and Cyrill Stachniss for providing the datasets.

\bibliographystyle{IEEEtran}
\bibliography{IEEEabrv,schaefer2018icra}

\end{document}